\documentclass[runningheads]{llncs}
\usepackage[T1]{fontenc}
\usepackage{graphicx}
\usepackage{amsmath,mathtools}
\usepackage{amssymb}

\begin{document}
\title{Generative Plant Growth Simulation from Sequence-Informed Environmental Conditions}
\titlerunning{SI-PGS: Sequence-Informed Plant Growth Simulation}
%
\author{Mohamed Debbagh\inst{1}\thanks{Corresponding author: mohamed.debbagh@mail.mcgill.ca}\orcidID{0009-0009-4636-1024} \and
Yixue Liu\inst{2}\orcidID{0000-0001-8316-5473} \and
Zhouzhou Zheng\inst{2}\orcidID{0000-0002-0822-1436} \and
Xintong Jiang\inst{1}\orcidID{0000-0003-4947-1981} \and
Shangpeng Sun\inst{1}\orcidID{0000-0001-7095-8626} \and
Mark Lefsrud\inst{1}\orcidID{0000-0002-7667-4269}}
\authorrunning{M. Debbagh et al.}
%
\institute{McGill University, Montréal Quebec H3A 0G4, Canada\\
\email{\{mohamed.debbagh,xintong.jiang\}@mail.mcgill.ca}\\ \email{\{shangpeng.sun,mark.lefsrud\}@mcgill.ca}
\and 
College of Mechanical and Electronic Engineering, Northwest A\&F University, Yangling Shaanxi 712100, China\\
\email{\{sunnyliu,zhengzz\}@nwafu.edu.cn}
}

\maketitle   
\begin{abstract}
A plant growth simulation can be characterized as a reconstructed visual representation of a plant or plant system. The phenotypic characteristics and plant structures are controlled by the scene environment and other contextual attributes. Considering the temporal dependencies and compounding effects of various factors on growth trajectories, we formulate a probabilistic approach to the simulation task by solving a frame synthesis and pattern recognition problem. We introduce a sequence-informed plant growth simulation framework (SI-PGS) that employs a conditional generative model to implicitly learn a distribution of possible plant representations within a dynamic scene from a fusion of low-dimensional temporal sensor and context data. Methods such as controlled latent sampling and recurrent output connections are used to improve coherence in the plant structures between frames of prediction. In this work, we demonstrate that SI-PGS is able to capture temporal dependencies and continuously generate realistic frames of plant growth.

\end{abstract}
\section{Introduction}
The phenotypic expression of a plant is significantly influenced by the dynamic growing conditions of its environment. In controlled environment agriculture (CEA), the goal is to control these conditions to produce desirable plant traits. As such, computer vision (CV) and sensor fusion-based phenomic models are commonly employed in CEA to assess crop quality and quantify biophysical properties throughout various stages of its growth cycle \cite{10.3389/fpls.2017.02233,doi:10.34133/plantphenomics.0080,10.3389/fpls.2018.00016}. More recently there has been an effort towards producing plant representation models that simulate plant structural development and predict biophysical attributes \cite{cieslak2022system,hitti2024growspace,prusinkiewicz2018modeling}. These simulations are often realized as parameterized functional-structural plant models (FSPM) that implement strict rules on physiological processes in the form of deterministic computational algorithms \cite{soualiou2021functional}. While FSPMs employ mechanisms that mimic plant processes and structural development patterns, they do not capture the true stochastic nature of plant growth and rely on fixed rendered representations of plant structures. In contrast, we propose a probabilistic approach to plant simulation through frame synthesis, a class of generative CV that aims to synthesize novel images of a scene captured by a distribution of observed data. In this paper, we outline a simulation framework that does not rely on rule-based processes to generate structures but rather a model that reflects a distribution of possible outcomes guided by an observation dataset in which the underlying physiological processes are implicitly learned. We summarize the contributions made in this work as follows:

\begin{itemize}
    \item We develop a sequence-informed generative model that performs a continuous mapping between low-dimensional context or "environment" data with high dimensional synthesized frames.
    \item We demonstrate that the stochastic elements that contribute to ambiguities in deterministic approaches can be controlled by learning a latent space to isolate variability in representations and maintain spatiotemporal consistency in plant structure development.
    \item We improve model performance by employing a recurrent connection from the output layers to capture dependencies from prior states.
    \item We evaluate the framework on perceptual quality and frame-by-frame coherence metrics along with qualitative observations. 
\end{itemize}

\section{Related Work}
\label{related_work}
Recent works have approached the task of plant growth modeling by representing structural elements through images and treatment data. Two prominent neural network (NN)-based approaches have been employed to predict future stages of plants' structural development from previous states found within a temporal scene.

\subsubsection{Next Frame Prediction}
In this approach, a frame-by-frame analysis using a convolutional recurrent neural network (CRNN) has been suggested as a method to extract features from a sequence of spatial image data \cite{keren2016convolutional}. Early works have taken a deterministic approach to the plant growth prediction problem by suggesting that the next sequence of a plant's structural development can be predicted by feeding a sequence of prior images through some implementations of a CRNN \cite{sakurai2019plant,yasrab2021predicting}. These methods use encoder-decoder networks to produce subsequent frames of the temporal scene. However, due to the ill-posed nature of the prediction task, there exist many possible outcomes for the plant's trajectory such that any deterministic generative process will result in ambiguities in outputs over large datasets \cite{kingma2013auto}. These ambiguities are further exacerbated by unguided predictions that provide no influence from scene conditions.

\subsubsection{Conditional Generative Forecasting}
Generative models have typically been employed as a data augmentation technique for training on sparse agricultural datasets \cite{doi:10.34133/plantphenomics.0041,lu2022generative}. There have been a few developments in plant growth forecasting which employ conditional generative models (CGM) for discrete predictions in which potential outcomes are generated at an arbitrary point in the future \cite{drees2021temporal,leonhardt2022probabilistic,Miranda_miro_9956115}. These predictions are often controlled by fixed treatment attributes passed as conditional inputs. Stochastic elements are introduced in the form of random noise to capture variability. However, these conditions are represented as discrete treatment embeddings and do not incorporate the dynamic and compounding environmental conditions that play a significant role in plant phenotype expression. Discrepancies in temporal consistencies in plant structures are observed when continuous predictions are made, as frames are not coherently coupled.

\begin{figure}[t]
\centering
\includegraphics[width=\textwidth]{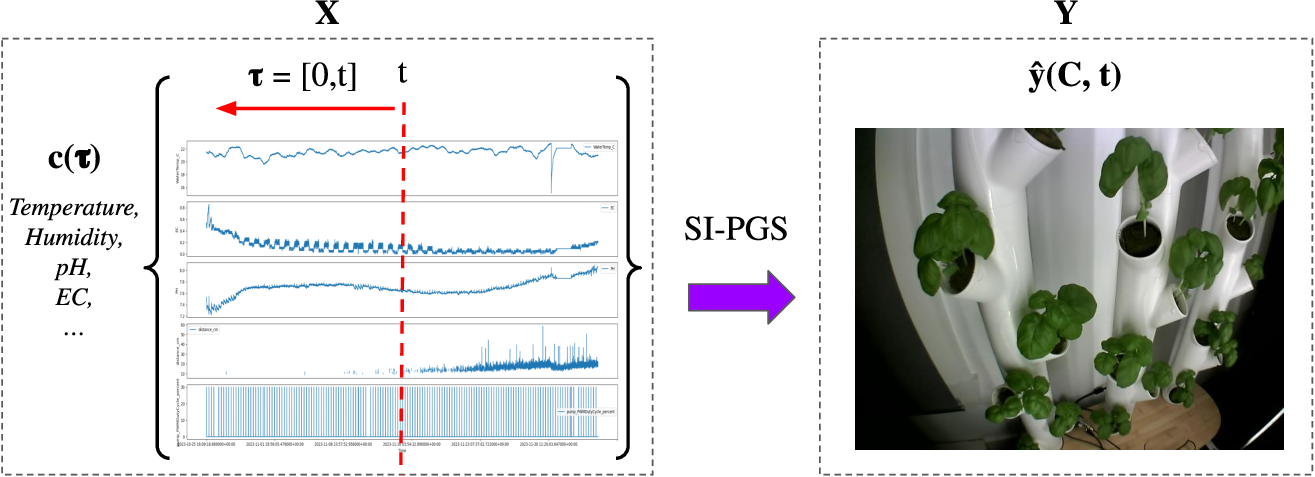}
\caption{Conceptual frame synthesis framework relating the $X$ sequence condition domain to the $Y$ structured output domain.} \label{frame_synth}
\end{figure} 

\section{Methods}
The task of plant simulation is formulated as a structured representation problem where we generate the frames of a scene, $y$, from a low-dimensional discrete or continuous random variable, $c$ (Fig. \ref{frame_synth}). The generated outputs, $\hat{y}$, of the model represent a simulated visualization of the plant's structural development and phenotypic attributes at a given point in time, $t$, with respect to its growth cycle. We construct a CGM framework, particularly a conditional variational auto-encoder (CVAE), with the objective of learning a prior distribution, $p_\theta(z|x)$. The latent representation of the scene, $z$, is directed by a set of encoded sequence vectors, $X = \{ x_i \mid i \in \{1,2, \ldots, N\}\}$, and is determined by the dataset of environmental conditions, denoted as $C=\{c_{i}(\tau) \mid i \in \{1, 2, \ldots, N\}, \tau \in [0,t]\}$ with $N$ i.i.d. samples. 

\begin{figure}[t]
\centering
\includegraphics[width=\textwidth]{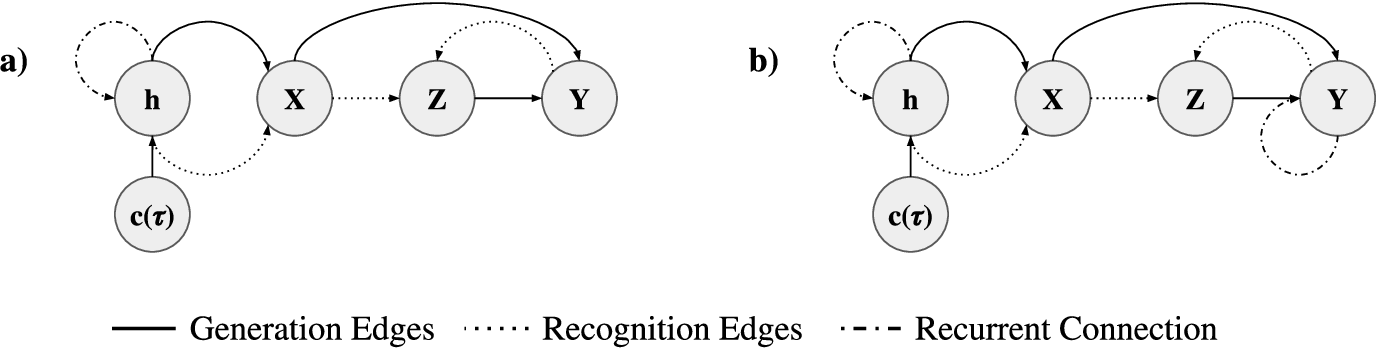}
\caption{Directed graph models. \textbf{(a)} SI-PGS. This model incorporates generation edges that represent the forward generative process and recognition edges that depict the encoding processes of the latent space, $Z$. The conditional vector, $X$, is influenced by the last layer hidden outputs, $h$, of an RNN that processes compounding sequential context data, $c(\tau)$, and is subsequently used to derive structured outputs, $Y$.
\textbf{(b)} SI-PGS with reccurent output connections (SI-PGS-R). Recurrent connections are introduced to the model to utilize feedback from the generated output $Y$ to refine $X$ and $Z$.}
\label{model overview}
\end{figure}

\begin{figure}[t]
\centering
\includegraphics[width=\textwidth]{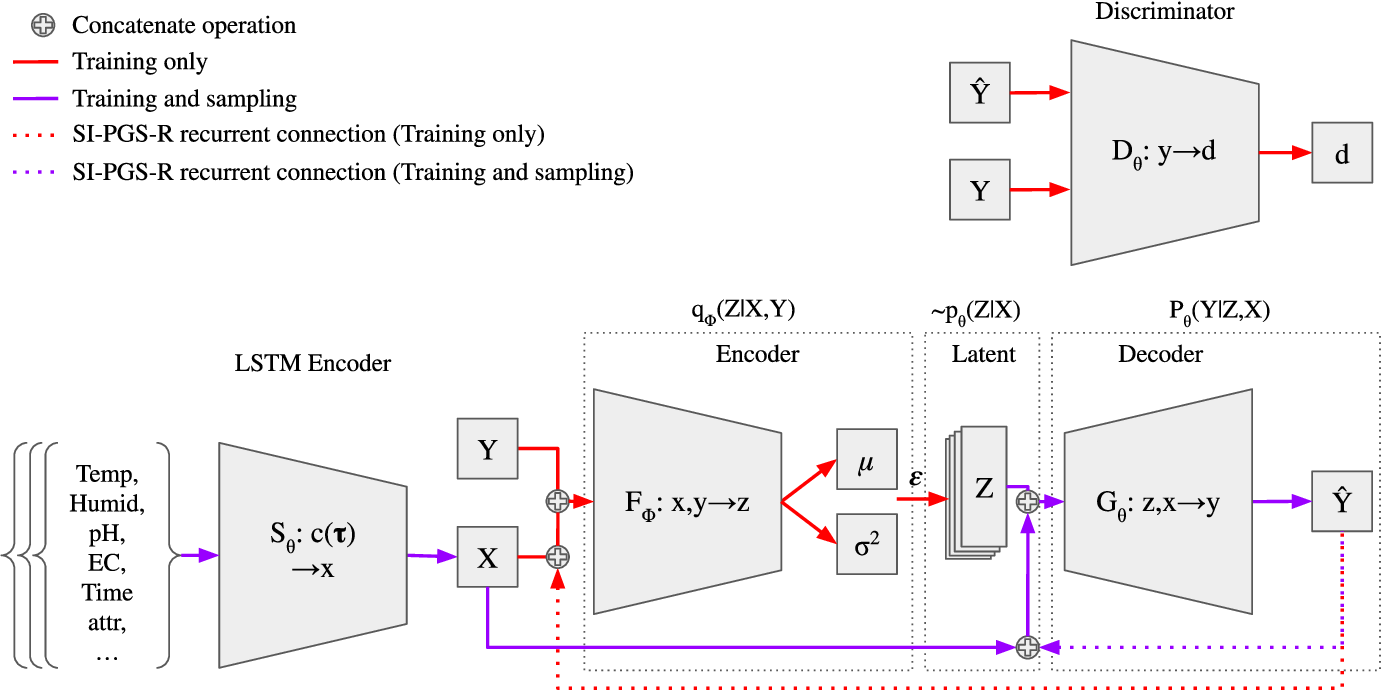}
\caption{SI-PGS network architecture.
Input, $c(\tau)$, is composed of a sequence of scene attributes (i.e. temperature, humidity, pH, EC and temporal attributes) that are encoded by the LSTM Encoder $S_\theta$.
The encoded output $x$ and the ground truth $Y$ are concatenated and fed into the encoder $F_\phi$, which produces the latent distribution variables $\mu$ and $\sigma^2$.
latent variable $z$ is sampled and passed through the decoder $G_\theta$ to generate the output $\hat{y}$.
The discriminator $D_\theta$ distinguishes between real and generated outputs. Implementation of the recurrent connection is denoted as SI-PGS-R.
} \label{Architecture}
\end{figure}

In this section, we outline the various components of the directed graph models and the implementation of our sequence-informed plant growth simulation (SI-PGS)(Fig. \ref{model overview}-\ref{Architecture}).

\subsection{Sequence-Informed Conditioning}
\label{Sequence informed conditioning}
$c$ is defined as a configuration of random variables that describe temporal observations contributing to the development and phenotypic expression of the plant within the scene. Typically, this is in the form of data observed from external sensors, intrinsic scene context, or discrete attributes. Accounting for the temporal aspect of our simulation, $c$ incorporates two time attributes with respect to the scene: The time elapsed since the inception of the plant's growth trajectory, defined at the first frame, and the time-interval between frames. Integrating these attributes allows us to relax our definition of $t$ to represent the sequence indices of each frame with variable time intervals and allows for the implicit interpolation of missing data between these periods. A sequence encoder, $S_\theta: c(\tau) \rightarrow x$, is employed to capture the accumulated effects of these observations on plant development at $t$.

\begin{equation}
\label{hidden state}
h_{t}=f_{\text{rnn}}(c(\tau), h_{t-1}; \theta) \quad \tau \in \{1, 2, \ldots, t\}
\end{equation}

\begin{equation}
\label{sequence embedding}
x = g_\theta(h_t)
\end{equation}

$S_\theta$ introduces an RNN, particularly a LSTM, denoted as $f_{\text{rnn}}(.)$, with parameters $\theta$, whose inputs represent compounding sequential information. The hidden states of the RNN, $h$, are mapped to a MLP forward process, $g_\theta(.)$ (Equation \ref{hidden state}-\ref{sequence embedding}).

\subsection{Latent Representation and Sample Generation}
\label{Latent representation and sample generation}
A prior distribution, $p_\theta(z|x)$, is learned so that in turn a distribution of generated scene outcomes, $\hat{y} \sim p_\theta(y|z,x)$, can be derived. The simulated outcome space, $Y$, is obtained by sampling the latent space, $Z$, such that $z \sim p_\theta(z|x)$ (Fig. \ref{model overview}a).

\subsubsection{Recognition Model}
The recognition model is given by the variational distribution, $q_\Phi(z|x,y)$. It is obtained implicitly by learning the distribution parameters, $\Phi$, through an encoding process, $F_\phi: x,y \rightarrow z$.

\begin{equation}
\label{CNN encoder}
\mu, \sigma^2 = f_\text{enc}(x,y; \Phi)
\end{equation}

\begin{equation}
\label{reparameterization trick}
z = \mu(x,y) + \sigma(x,y) \odot \epsilon
\end{equation}

We define $q_\Phi(z|x,y)$ as a Gaussian distribution with $\Phi$ consisting of mean, $\mu$, and standard deviation, $\sigma^2$, such that $\mathcal{N}(\mu(x,y), \sigma^{2}(x,y))$. $f_\text{enc}(.)$ describes the convolutional encoding network which outputs a vector representation of the distribution parameters in Equation \ref{CNN encoder}. $Z$ can then be sampled by defining a deterministic functional parameterized by $\Phi(x,y)$. In order to perform backpropagation a stochastic auxiliary variable, $\epsilon \sim \mathcal{N}(0, I)$, is introduced to the functional and is defined as the reparameterization trick in Equation \ref{reparameterization trick} \cite{kingma2013auto}.

\subsubsection{Generation Model}
The generation model is outlined by, $G_\theta: z,x \rightarrow y$, the process in which a frame of the scene is synthesized such that $y \sim p_\theta(y|x,z)$. In our implementation of SI-PGS we employ a condition-dependent parameterization to achieve a sampling process that describes $z \sim p_\theta(z|x)$. First we obtain $z$ as a sample from a fixed Gaussian distribution $\mathcal{N}(0, I)$. Then, we introduce $x$ as the conditioning term that will modulate the parameters during generation. Both terms are passed as input to a convolutional decoder network, $f_{\text{gen}}(.)$, in Equation \ref{CNN decoder} to produce a constrained output.

\begin{equation}
\label{CNN decoder}
\hat{y} = f_{\text{gen}}(z,x;\theta) \mid z \sim \mathcal{N}(0, I)
\end{equation}

The recognition and generation models are characterized by an objective function which aims to optimize the variational lower bound, $\mathcal{L}(\theta, \Phi)$ \cite{NIPS2015_8d55a249}. The lower bound can be expressed by two terms; the first term, $D_{\text{KL}}(q_{\phi} \| p_{\theta})$, is the Kullback–Leibler ($D_{\text{KL}}$) divergence between the approximate posterior and the prior distributions. The second term, $\mathbb{E}_{q_{\phi}(z|x,y)}[p_{\theta}(y|x,z)]$, is the expected value of the log-likelihood. These terms act to balance the model between accurate reconstruction of the output, $\hat{y}$ with $y$, and a regularization on the learned latent space, such that the $q_{\Phi}(z|x,y)$ remains close to the prior $p_\theta(z|x)$. In previous work \cite{debbagh2023learning}, we demonstrate that the parameterization of $\mathcal{L}(\theta, \Phi)$ can be done to prioritize model learning between reconstruction and regularization. We employ a weight, $\beta$, in Equation \ref{cvae of} which introduces a tunable parameter that balances the importance placed by each term during training. $\beta>0.5$ signifies more importance placed towards the regularization term, while $\beta<0.5$ places more significance towards the reconstruction term.

\begin{equation}
\mathcal{L}_{\text{cvae}}(\theta, \Phi)  = - (\beta) D_{\text{KL}}(q_{\phi}(z|x,y) \| p_{\theta}(z|x)) + (1-\beta) \mathbb{E}_{q_{\phi}(z|x,y)}[p_{\theta}(y|x,z)]
\label{cvae of}
\end{equation}

\subsubsection{Discriminator}
$G_{\theta}(.)$ can be interpreted as the generator of a generative adversarial network (GAN) which employs a predefined random noise distribution, $\epsilon$, to guide image generation \cite{NIPS2014_5ca3e9b1}. As such, we can implement a discriminator, $D_{\theta}: y \rightarrow d$, whose output, $d$, represents the probability that input, $y$, is real or fake. SI-PGS employs a convolutional encoder network, $f_{\text{discr}}(.)$, whose output features are passed through a sigmoid function to quantify $d$ as shown in Equation \ref{discriminator}. This GAN mechanism is employed to refine the realistic generation of the synthesized frames by optimizing the value function, $V$, of a min-max game between the generator and discriminator with the objective function described in Equation \ref{GAN OF}. The combined SI-PGS objective function is given by Equation \ref{sipgs OF}.

\begin{equation}
\label{discriminator}
 d = \text{SIGMOID}(f_{\text{discr}}(y;\theta))
\end{equation}

\begin{equation}
\label{GAN OF}
\min_G \max_D V(D, G) = \mathbb{E}_{y \sim p(y)}[\log D(y|x)] + \mathbb{E}_{z \sim q_\phi(z|x,y)}[\log (1 - D(G(z|x)))]
\end{equation}

\begin{equation}
\label{sipgs OF}
\mathcal{L}_{\text{SI-PGS}} = \mathcal{L}_{\text{CVAE}}(\theta, \phi) + \min_{G} \max_{D} V(D, G)
\end{equation}

\subsection{Recurrent Connection}
\label{SI-PGS-R}
 A recurrent connection is incorporated into the architecture such that output, $\hat{y_t}$, is conditioned on the previous output state, $\hat{y}_{t-1}$ (Fig. \ref{model overview}b). The aim of this recurrent step is to introduce a known spatial state that will add continuity over time, resulting in improved frame generation. Thus the generation model is updated to $G_\theta: z,x,y_{t-1} \rightarrow y_t$, where $f_{\text{gen-R}}(.)$ represents the modified convolutional decoder network in Equation \ref{CNN decoder -R}. The implementation of the recurrent connections is shown in Fig. \ref{Architecture}.

\begin{equation}
\label{CNN decoder -R}
\hat{y_t} = f_{\text{gen-R}}(z,x,\hat{y}_{t-1};\theta) \mid z \sim \mathcal{N}(0, I)
\end{equation}

\section{Experiments}
In this section, we evaluate the various components of the simulation model. The objective of these experiments is to demonstrate that SI-PGS is able to synthesize continuous frames that capture the realistic development of a plant given a sequence of environmental data and outperforms the existing baseline model.

\subsection{Gardyn Dataset}
The Gardyn dataset (GD) is comprised of images and sensor data that observe Genovese basil (\textit{Ocimum basilicum}) grown in a controlled vertical hydroponics system. The data was collected over four months, covering two growth cycles and is used as a preliminary dataset for model testing. It contains time-series data which include 40,000 640 by 480 pixel RGB images, captured at two angles of the scene, each with corresponding sensor data taken at a frequency of 10 minutes. The sensor data include the following information: ambient temperature, ambient humidity, reservoir level, reservoir temperature, nutrient pH, and electrical conductivity (EC). GD includes metadata such as lighting schedule, irrigation schedule, timestamp information and image metadata. 

\subsection{Quantitative Model Evaluation}
\subsubsection{Frame Generation}
One major criterion for evaluating the generative component of SI-PGS is the realistic synthesis of scene frames. As such, two metrics are selected to quantitatively assess the generated outputs: reconstruction loss and Fréchet inception distance (FID) score.

A reconstruction loss, particularly average mean squared error (MSE), is employed to quantify the pixel-wise discrepancies between the generated frames and their corresponding real images from GD. This metric measures the overall visual fidelity of the target scene frame. The FID score evaluates the distance between feature vectors calculated for real and generated images, which are extracted using an Inception-v3 model pre-trained on ImageNet \cite{Szegedy_7780677}. This metric considers both the feature mean and covariance, capturing not just the visual appearance but the statistical properties of the image distributions. A lower FID score indicates that the generated images are not only visually similar to the actual images but are statistically comparable, reflecting better perceptual quality in the synthesized frames \cite{Heusel_NIPS2017_8a1d6947}.

\subsubsection{Coherence Between Frames}
Temporal coherence, within the bounds of our simulation, is defined by the smooth and proportional development of plant structures between frames with respect to time. In other words, objects within a scene should not drastically change from one frame to the next if the time increments are small. These temporal inconsistencies are referred to as "jitters". We employ two coherence metrics: Time-weighted MSE and structural similarity index (SSIM). These measures are used to compare differences in sequential frames of predictions. 

Since MSE emphasize larger errors and is sensitive to outliers and abrupt changes between frames, it is employed to assess the stability of frame transitions. SSIM, measures the similarity between two images by assessing the visual impact of three characteristics of an image: luminance, contrast, and structure. This metric provides a more comprehensive measure of similarity by considering how changes in the structural information and luminance affect the perceived quality of video sequences \cite{wang2004video}. SSIM is measured between 0 and 1, with more similarities between images as the value approaches 1. Since the frames synthesized by SI-PGS are not necessarily generated at fixed time intervals, we extend these metrics by applying a weighted time parameter, $w_t$, which accounts for the time interval between generated frames, normalized across all weights.

\subsection{Experimental Setup}
We evaluate the performance of the SI-PGS framework compared with a baseline discrete embedding conditional GAN (cGAN), conceptually used in preceding work for predictive plant growth modeling \cite{Miranda_miro_9956115}. The cGAN model is deployed with a comparable network architecture and depth, which is obtained by detaching the generator and discriminator from the latent encoding process of SI-PGS and using the objective function given by Equation \ref{GAN OF}. A modified SI-PGS with recurrent output connections, outlined in Section \ref{SI-PGS-R}, is tested alongside the original model and is denoted as SI-PGS-R. We experiment with various $\beta$ values and observe the effect of reconstruction versus regularization on the representation of plant structures during simulation. The $\beta$ values selected for the experiments are: 0.10, 0.25, 0.50, 0.75 and 0.90. We further test a controlled and fixed latent space sampling process, denoted as controlled sampling (CS), to isolate the stochastic elements that lead to variability in consecutive frames. The models are trained on a subset of GD containing 5,892 images and corresponding sequence data. Moreover, all models are trained for 70 epochs, with early stopping to avoid overfitting once the model converges on the validation set. The experiments in this study were conducted using a workstation equipped with an NVIDIA GeForce RTX 4080 GPU. The full implementation and supplemental materials from these experiments are available on GitHub.\footnote{Code and supplementary material can be found at: \\ \url{https://github.com/mohas95/Sequence-Informed-Plant-Growth-Simulation}}

\begin{table}[t]
    \caption{Comparative analysis of model performance evaluating generation quality and frame coherence metrics. Time-weight (tw) applied to frame coherence metrics follows an exponential decay, $w_t = e^{-\lambda \cdot \Delta \text{t}}$, with a decay factor, $\lambda = 1.9 \times 10^{-4}$, and is normalized over $\sum_{t=2}^{n-1}w_t$.} \label{tab1}
    \centering
    \begin{tabular}{c | c c | c c  c c}
    \hline
    &\multicolumn{2}{c|}{\bfseries Generation Quality}& \multicolumn{4}{c}{\bfseries Frame Coherence} \\
    
    & FID & MSE & $\text{MSE}_{\text{tw}}$ & $\text{SSIM}_{\text{tw}}$ & $\text{CS-MSE}_{\text{tw}}$ & $\text{CS-SSIM}_{\text{tw}}$\\
    \hline
    SI-PGS ($\beta=0.25$)& 142.78 & 8915 & 2942  & 0.844  & 1.08 & $\geq0.999$\\
    SI-PGS ($\beta=0.50$)& 88.86 &  6359 & 2081 & 0.904 & 1.16  & $\geq0.999$\\
    SI-PGS ($\beta=0.75$)& 62.25 & 3389 &  1279 & 0.937 & 1.04 & $\geq0.999$\\
    SI-PGS ($\beta=0.90$) & 47.04  &  \bfseries 2441 & \bfseries1024  & \bfseries 0.951 & \bfseries 0.89 & $\geq0.999$\\
    \hline
    SI-PGS-R & \bfseries 45.09 & 6858 & 6144 & 0.889 & 5.98 & $\geq0.999$\\
    \hline
    cGAN & 218.33 & 14002 & 3365 & 0.930 & $\leq 0.001$ & $\geq0.999$ \\
    \hline

    \end{tabular}
\end{table}

\section{Results and Discussion}

\begin{figure}[t]
\centering
\includegraphics[width=\textwidth]{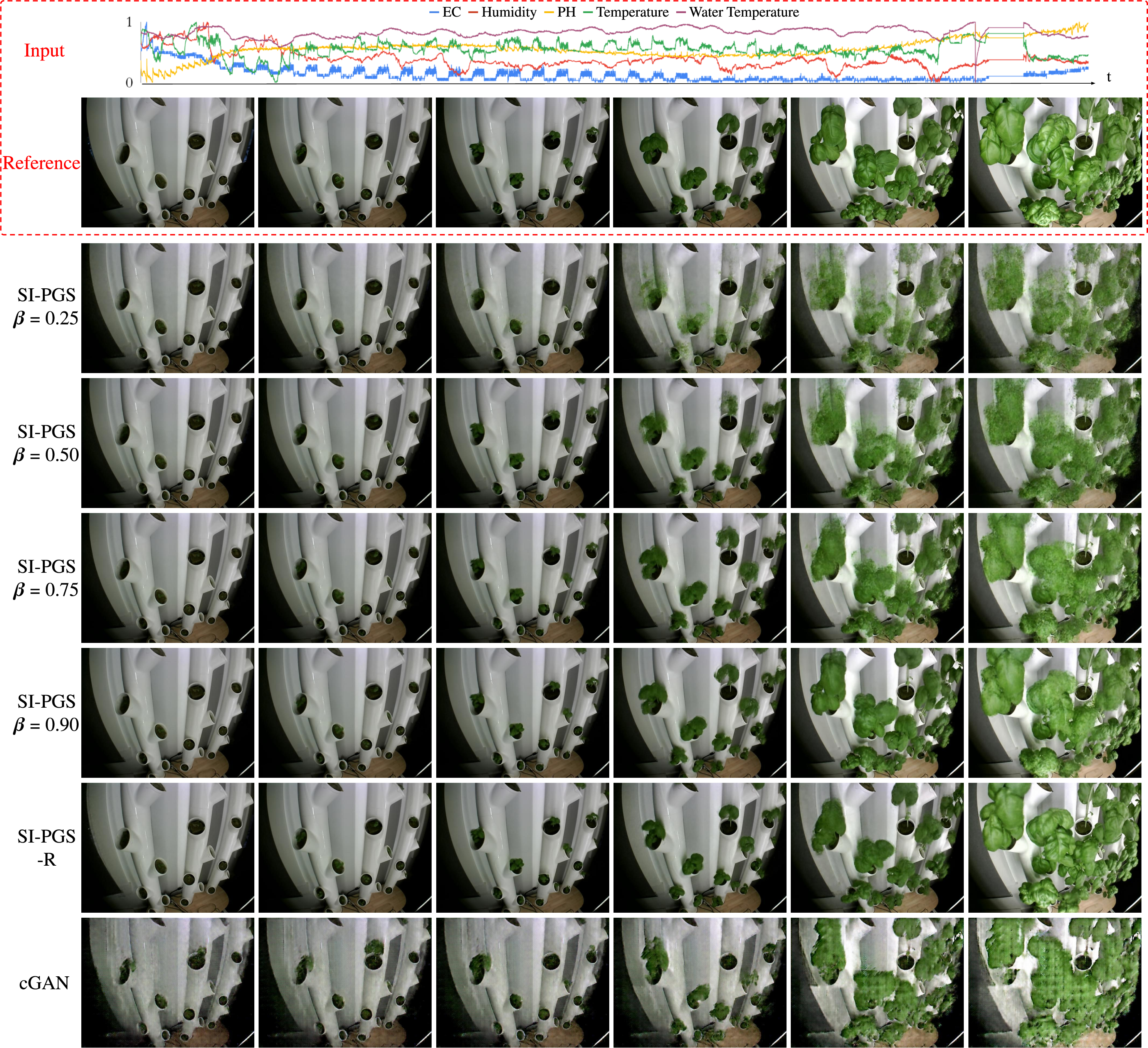}
\caption{Comparative generated model outputs. Top two rows show the input signals over time and corresponding real reference frames. Subsequent rows show the outputs of corresponding generative models.} \label{results}
\end{figure}

The results from the quantitative experiments are shown in Table \ref{tab1} and demonstrate that SI-PGS outperforms cGAN in generating sequence-consistent plant structures. FID scores remain consistently lower for SI-PGS based models, with the lowest obtained from SI-PGS-R, indicating more apparent similarities in identifiable structures between generated and reference images. When fixing the stochastic elements of each model ($z$ for SI-PGS and $\epsilon$ for cGAN) we observe a significant decrease in CS-$\text{MSE}_{tw}$ while CS-$\text{SSIM}_{tw}$ approaches 1 across all models suggesting that jitters between frames are suppressed. This is interpreted as the effective isolation of variability that leads to ambiguities in deterministic models. Thus, each latent variable can be used to produce a unique instance of a scene outcome with spatiotemporal consistent structural development. Finally, we notice a consistent decrease in FID score and MSE with increasing $\beta$ values. This suggests that the outputs of the models that favor the regularization term over the reconstruction in the objective function produce more realistic structures. This is more apparent with the qualitative assessment of model outputs. 

\subsection{Qualitative Assessment}
Outputs of the experiment are shown in Fig. \ref{results} and illustrate the sequential frames of a scene generated from low-dimensional scene attributes. SI-PGS models are able to capture plant structures across all references, while the cGAN model fails to capture early stages of the plant. Moreover, we suggest that cGAN requires a longer training phase to generate realistic plant structures when compared to SI-PGS, as later-stage frames fail to capture finer plant details.

\subsubsection{Regularization vs Reconstruction}
We observe that as the $\beta$ value increases, the outputs of the model become more realistic when favoring the regularization term. While intuitively the reconstruction term aims to increase 
the model's pixel-wise fidelity, in reality, it results in an averaging effect across large image datasets. In effect, the reconstruction term favors a deterministic outcome, leading to the ambiguous structures in generated outputs.

\subsubsection{Effects of Recurrent Connection}
While the effects of the SI-PGS-R are not captured well in Fig. \ref{results}, it becomes apparent when observing all frames compiled into video format. The addition of a recurrent connection from the output layers of the model results in a feedback mechanism that enhances the model's ability to capture dependencies and improve generative performance. The generated frames capture movements of the plant structure during growth more realistically over the standard SI-PGS model.

\section{Conclusion}
In this work, we propose a novel generative framework for plant growth simulation that captures continuous temporal dependencies. We demonstrate its effectiveness in reconstructing plant structural development using a typical CEA image dataset and compare it to a preceding benchmark generative approach. Furthermore, we introduced variations to the baseline model to improve realistic and coherent scene synthesis. Additional research should be done to link phenotypic information to synthesized frames. Such efforts should be directed towards building larger datasets that capture the temporal dependencies in plant structural development and phenotypic attributes.

\subsubsection{Acknowledgements}
This study was partially funded by Gardyn and Mitacs (IT16220). We thank the Gardyn team for providing the vertical growth systems essential for building our datasets. Special thanks to Anita Parmar, Ollivier Dyens, and the Building 21 members for their engagement in creative discussions.

\bibliographystyle{splncs04}
\bibliography{main}

\end{document}